\documentclass[sigplan,nonacm]{acmart}

\AtBeginDocument{%
  }

\setcopyright{none}
\renewcommand\footnotetextcopyrightpermission[1]{}

\usepackage{multirow}
\usepackage{subfigure}
\usepackage{bbding}
\usepackage{makecell}
\usepackage{mathtools}
\usepackage[super]{nth}
\usepackage{url}

\usepackage{algorithm}
\usepackage{algpseudocode}
\algtext*{EndFor}
\algtext*{EndFunction}
\algtext*{Until}
\algtext*{EndIf}
\algtext*{EndWhile}
\usepackage{amsmath}
\usepackage{array}
\usepackage{tablefootnote}

\usepackage{geometry}

\usepackage{soul}
\usepackage{siunitx}
\usepackage{multicol}

\newcommand{\sys}{AgentZip\xspace}

\newcommand{\yao}[1]{\textcolor{black}{#1}}

\usepackage{xspace}
\usepackage{tikz}
\newcommand*\circled[1]{%
  \tikz[baseline=(char.base)]{%
    \node[shape=circle, draw, inner sep=0.8pt] (char) {#1};%
}%
}

\begin{document}

\title{Memory Compression for High-Fanout Agent Sandboxes}

\author{Mengming Li}
\authornote{Mengming Li and Ceyu Xu contributed equally to this work.}
\email{mengming.li@connect.ust.hk}
\affiliation{%
  \institution{HKUST}
  \country{}
}

\author{Ceyu Xu}
\authornotemark[1]
\email{eeentropy@ust.hk}
\affiliation{%
  \institution{HKUST}
  \country{}
}

\author{Qijun Zhang}
\email{qzhangcs@connect.ust.hk}
\affiliation{%
  \institution{HKUST}
  \country{}
}

\author{Jiangnan Yu}
\email{jyucr@connect.ust.hk}
\affiliation{%
  \institution{HKUST}
  \country{}
}

\author{Xiangfeng Sun}
\email{xsunbv@connect.ust.hk}
\affiliation{%
  \institution{HKUST}
  \country{}
}

\author{Haohui Mai}
\email{Haohui@ust.hk}
\affiliation{%
  \institution{HKUST}
  \country{}
}

\author{Zhiyao Xie}
\authornote{Corresponding author.}
\email{eezhiyao@ust.hk}
\affiliation{%
  \institution{HKUST}
  \country{}
}

\begin{abstract}

High-fanout agent workloads create a growing memory bottleneck because a single task may spawn many concurrent sandbox sessions. Yet these sandboxes are far from independent: they originate from a shared template and execute related trajectories, exposing substantial template-relative and cross-sandbox memory redundancy. Conventional memory compression is poorly matched to this setting in three fundamental dimensions: \emph{how} to compress, because they fail to exploit similarity across non-identical sandbox pages; \emph{what} to compress, because they control page-fault overhead through conservative page selection; and \emph{when} to compress, because compression is either triggered by memory pressure or performed without awareness of agent execution phases.

We present AgentZip, the first memory compression system designed specifically for AI-agent sandboxes. AgentZip introduces compression mechanisms that exploit both the template-relative and cross-sandbox redundancy. It broadens the compression scope to any page with a profitable representation and shifts overhead control from compression-time page selection to restore-time prefetching. It further aligns expensive compression with LLM waiting periods to avoid interfering with foreground tool execution. Across LLM training and inference workloads, AgentZip reduces sandbox-owned memory by up to 8.7$\times$, compared with 2.1$\times$ for the Linux configuration. Restore prefetching and agent-execution-aware scheduling reduce the slowdown of aggressive compression from as high as 3.1$\times$ to 1.40$\times$ while retaining nearly all of its memory-saving benefit.

\end{abstract}

\maketitle
\pagestyle{plain}

\section{Introduction}
\label{sec:intro}

\begin{figure}[t]
\centering
\includegraphics[width=.99\linewidth]{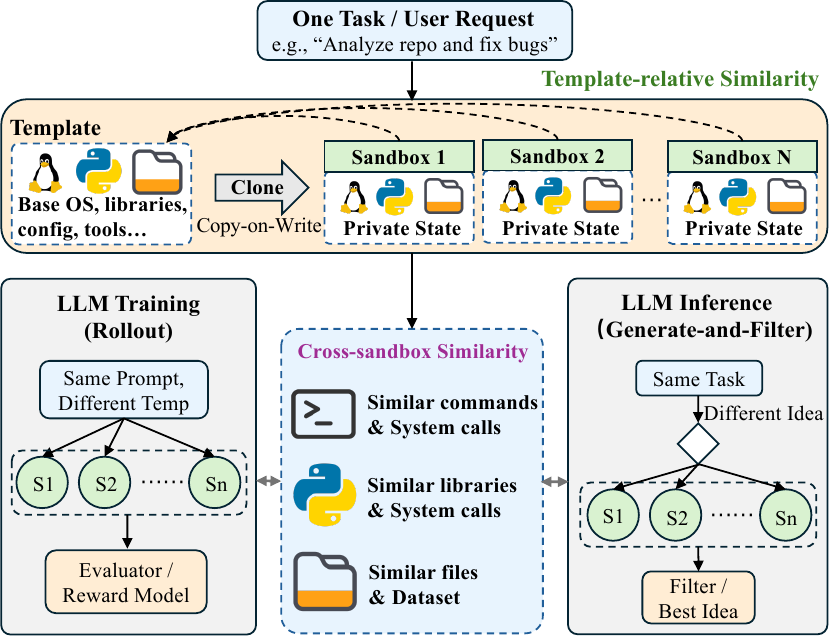}
\caption{Similarity between sandboxes creates opportunities for inter-sandbox memory compression. }
\vspace{-.2in}
\label{fig:similarity}
\end{figure}

Modern AI agents no longer merely produce text. To resolve a software issue,
analyze a dataset, or operate a computer, an agent typically issues shell commands, edits
files, installs dependencies, and runs
tests~\cite{wang2025openhands,yang2024swe,jimenez2024swe}. Because these actions
execute untrusted code and mutate persistent state, agent platforms nowadays
choose to run every action inside a sandboxed execution environment that
provides isolation and lifecycle
control~\cite{young2019gvisor,agache2020firecracker}. As a result, the agent
sandbox is no longer a peripheral part of the serving stack: it is where agent
execution actually happens.

\textbf{Memory as bottleneck due to scaling.} Agentic workloads are increasingly
\emph{high-fanout}: a single task spawns many sandboxes rather than one. 
As shown in Figure~\ref{fig:similarity}, during reinforcement learning (RL) \emph{training}, an agent samples tens of independent trajectories per
task so that an RL reward model can score them~\cite{sheng2025hybridflow,zhang2026prorlagent}. 
During \emph{inference}, multi-agent
workflows run several candidate sessions concurrently to speed up parallel trial-and-errors~\cite{ni2026chimera}. 
In summary, fanout enables productive LLM training and inference, but it also makes them expensive. Specifically, it changes the unit of provisioning: a host no longer backs one sandbox per task, but tens of concurrent sibling sandboxes. Crucially, the resource that usually runs out first is \emph{memory} instead of \emph{compute}. 
Sandboxes spend much of their life blocked on model output: between two tool calls, a sandbox sits idle waiting for the LLM to decode the next command, so cores stay available while memory does not. 

\textbf{Opportunity: Redundancy in Memory.} 
If each sandbox occupies $M_s$ bytes and a host devotes \yao{altogether} $M_h$ bytes to sandboxes, concurrency is capped near $M_h/M_s$. \yao{However,} this linear scaling is far worse than the workload deserves. Sibling \yao{sandboxes} that fan out from one task are not independent: they are cloned from the same template image, open the same repository, import the same libraries, and run overlapping command sequences. Their memory redundancy comes from two complementary sources\yao{:} \yao{1)} \emph{Template-relative similarity}\yao{. It} arises because sandboxes are cloned from a shared template, and many private pages remain close to their corresponding template pages after copy-on-write. However, this similarity can weaken as execution progresses and sandboxes accumulate more private state\yao{s}. \yao{2)} \emph{Cross-sandbox similarity}\yao{. It} provides a complementary source of redundancy: sibling sandboxes executing related trajectories often develop similar runtime states even after they have diverged from the original template. Section~\ref{sec:eval-similarity} shows both forms of similarity are substantial, with 76--96\% of pages exhibiting template-relative or cross-sandbox redundancy in our measurements.

At a high level, a memory compression system must answer three fundamental questions: \yao{1)} \emph{How to compress}, which determines which compression algorithms and representations are used to exploit available redundancy. \yao{2)} \emph{What to compress}, which determines which memory pages are selected as candidates for compression. \yao{3)} \emph{When to compress}, which determines when compression is performed and physical pages are reclaimed. \yao{We point out that all these three fundamental decisions in} today's mainstream memory compression mechanisms are poorly matched to high-fanout agent sandboxes\yao{, as we elaborate below:}


\circled{1} \textbf{[How]: Wrong redundancy model.} General-purpose compressors such as zswap\yao{~\cite{zswap-doc,zswap-lwn}} and zram~\cite{linux-zram} \yao{only} exploit \emph{intra-page} redundancy. \yao{These general compressors} compress each page independently and therefore cannot exploit its similarity to the original template or to runtime content in sibling sandboxes. Page-level deduplication such as KSM~\cite{linux-ksm} exploits \emph{inter-page} redundancy, \yao{but requires exact page equality. Most of such exact-sharing opportunities have already been} covered by copy-on-write at sandbox creation. 
\circled{2} \textbf{[What]: Wrong compression scope.} Existing compression systems~\cite{panwar2022tmcc,lagarcavilla2019software,kumar2026tierscape} restrict compression primarily to pages unlikely to be accessed in the near future, using access hotness or predicted reuse to control page-fault overhead. This creates an inherent trade-off: compressing \yao{warmer} pages exposes greater memory-saving opportunities, but also increases costly page faults. Accurately identifying which warm pages can be compressed is difficult. Thus, conservative selection \yao{becomes inevitable, leaving} many profitable compression opportunities \yao{wasted}. 
\circled{3} \textbf{[When]: Wrong timing.} Zswap\yao{~\cite{zswap-doc,zswap-lwn}} operates only reactively on the swap-out path and under memory pressure. This compression timing is poorly matched to reducing the \emph{average} sandbox memory footprint over its lifetime, because compressible pages may remain resident for most of a session.

This paper presents \sys, a sandbox memory compression system that rethinks these three \yao{fundamental} design choices for high-fanout agent workloads. 

\textbf{[How]: Sandbox-aware redundancy model.} \sys directly exploits the two forms of memory redundancy exposed by sandboxes. \yao{1)} For \emph{template-relative} similarity, it represents a sandbox's private page as a compact delta from its corresponding immutable template page.  \yao{2)} For \emph{cross-sandbox} similarity, it builds shared dictionaries that capture common byte patterns across sibling sandboxes and uses these dictionaries to encode candidate pages. These mechanisms broaden the exploitable inter-page redundancy beyond exact page equality. \sys further combines them with lightweight intra-page compression and selects the most space-efficient representation for each page.

\textbf{[What]: Unlimited compression scope.} \sys no longer restricts compression eligibility based on page hotness or predicted reuse. \yao{Instead,} any page is eligible for compression as long as it can be compressed profitably, including warm pages that are likely to be accessed again. This broader scope is enabled by a \emph{new compression paradigm}: \yao{Instead of} avoiding future page faults through selection at \emph{compression time}, \sys \yao{allows compressing warm pages by mitigating page faults} at page \emph{restore time} through prefetching. \yao{At restore time, \sys} predicts which compressed pages will be needed soon and prefetches them before demand.


\textbf{[When]: Agent-execution-aware compression timing.} Rather than waiting \yao{until} memory pressure \yao{to start compression}, \sys \yao{proactively} aligns compression with the agent execution \yao{lifecycle}: it performs lightweight candidate discovery during \emph{tool execution} and moves expensive compression into \emph{LLM waiting periods}. This lifecycle awareness avoids interfering with foreground tool execution while reducing the time-averaged sandbox memory footprint.




\yao{The contributions of this paper are summarized below:} 
\vspace{-.1in}
\begin{itemize}
  \item We present \sys, the first memory compression system that exploit the redundancy within AI-Agent sandboxes. We identify sandbox memory as a first-order scaling bottleneck in high-fanout agent workloads and show that sandbox-specific redundancy and execution structure create substantial opportunities beyond conventional memory compression.
  \item We design a sandbox-aware compression algorithm that exploits both template-relative and cross-sandbox redundancy, with lightweight local compression. By capturing redundancy that existing mechanisms miss, \sys reduces sandbox-owned memory by up to 8.7$\times$ and four times more than Linux configuration.
  \item We rethink how compression overhead is controlled. \sys decouples compression from future page faults by aggressively compressing profitable pages and predicting their restores before demand, while aligning expensive compression with repeated LLM waiting periods rather than memory pressure. These techniques reduce the slowdown of aggressive compression from 3.1$\times$ to 1.40$\times$ while retaining nearly all of its memory-saving benefit.
\end{itemize}

\section{Background}



\subsection{Memory Compression}
\label{subsec:back_memcomp}

Memory compression increases effective memory capacity by replacing resident pages with smaller compressed representations stored in a memory pool. After a page is compressed, its original physical frame can be released, and only the compressed representation remains in memory. If the application accesses this page again, the system must first retrieve and decompress it from the compression pool and restore the original page before execution can continue. Therefore, memory compression \yao{essentially trades \emph{restore overhead} for \emph{memory capacity}}. 
The effectiveness of a memory compression scheme is jointly determined by \yao{3 main factors: compression \emph{redundancy model}, \emph{scope}, and \emph{timing}.}

\subsection{Compression Redundancy Model} 
\label{subsec:comp}
Existing Linux memory-saving mechanisms exploit redundancy using two representative approaches: \emph{page deduplication} and \emph{page-local compression}. Kernel Samepage Merging (KSM)~\cite{linux-ksm} is a mature Linux deduplication mechanism originally developed for KVM virtualization. A background kernel thread scans anonymous pages and replaces pages with identical contents by a single write-protected physical page. Subsequent writes trigger copy-on-write and create a private copy. KSM therefore exploits \emph{inter-page} redundancy, but requires exact page equality.

Zswap~\cite{zswap-doc,zswap-lwn} takes a different approach. When Linux swaps out an anonymous page, zswap compresses the page and stores its compressed representation in a memory-resident pool instead of immediately writing it to the backing swap device. The original physical frame can then be reclaimed. If the page is accessed again while its compressed copy remains in the pool, zswap retrieves and decompresses it during swap-in. Zswap supports compressors such as Zstd, LZ4, and LZO, to exploit byte-level redundancy \emph{within each page}.

\textbf{Limitations.} KSM captures \emph{inter-page} redundancy only when two pages are exactly identical, while much of this exact sharing is already provided by copy-on-write at sandbox creation. Zswap compresses each page independently and exploits \emph{intra-page} redundancy. Neither mechanism \yao{can capture} the inter-page approximate similarity between a private page and its template or among pages in sibling sandboxes.


\subsection{Compression Scope}
\label{subsec:scope}

Memory compression can introduce significant performance overhead when a compressed page is accessed again, because the page must be restored before execution can continue. Existing memory compression systems commonly control this overhead through \emph{page selection}, deciding which pages are safe to compress based on their expected future accesses. Prior systems~\cite{panwar2022tmcc,lagarcavilla2019software,kumar2026tierscape} typically formulate this as \emph{cold-page identification} and prioritize pages that are unlikely to be accessed in the near future. However, prior studies show that only about 20--30\% of memory pages are cold, while another 50--60\% are warm (i.e., not continuously accessed but still likely to be revisited after a moderate interval~\cite{kumar2026tierscape}).


\textbf{Limitations.} Existing hotness-based policies \yao{will severely limit the full exploitation of inter-sandbox redundancy.} Even with perfect cold-page identification, a cold-only policy leaves many potentially compressible warm pages untouched. Expanding compression to warm pages exposes more memory-saving opportunities, but also increases page-restore overhead. \yao{Prior study~\cite{kumar2026tierscape}} reports that extending compression from cold to cold-plus-warm pages increases memory saving from 11\% to 32\%, while slowdown \yao{significantly} rises from 9.5\% to 20\%. This trade-off makes it difficult for selection-based policies to aggressively expand the compression scope.

\subsection{Compression Timing} 
\label{subsec:timing}

Memory compression systems must also decide when compression should be performed. Many OS-level compressed-memory systems, such as zswap~\cite{zswap-doc,zswap-lwn}, operate \emph{reactively}: compression occurs only after memory reclaim selects a page for eviction and sends it down the swap-out path. This design integrates naturally with existing memory reclamation, but compression begins only when memory pressure exposes pages to reclaim. More recent systems\yao{~\cite{lagar2019software,weiner2022tmo}} instead perform compression \emph{proactively}, identifying and compressing candidate pages before they would otherwise be evicted.

\textbf{Limitations.} Neither reactive nor generic proactive timing is suitable for agent sandboxes. Reactive compression can act too late for reducing the \emph{time-averaged} sandbox memory footprint, because compressible pages may remain resident for most of a session before memory pressure occurs. Generic proactive compression exposes more opportunities, but it is unaware of the sandbox execution lifecycle and may perform expensive compression while a tool call is running, directly interfering with foreground execution. Agent sandboxes therefore require compression timing that is both proactive and coordinated with their execution phases.

\section{Overview}
\label{sec:software}

\begin{figure*}[t]
\centering
\includegraphics[width=.99\linewidth]{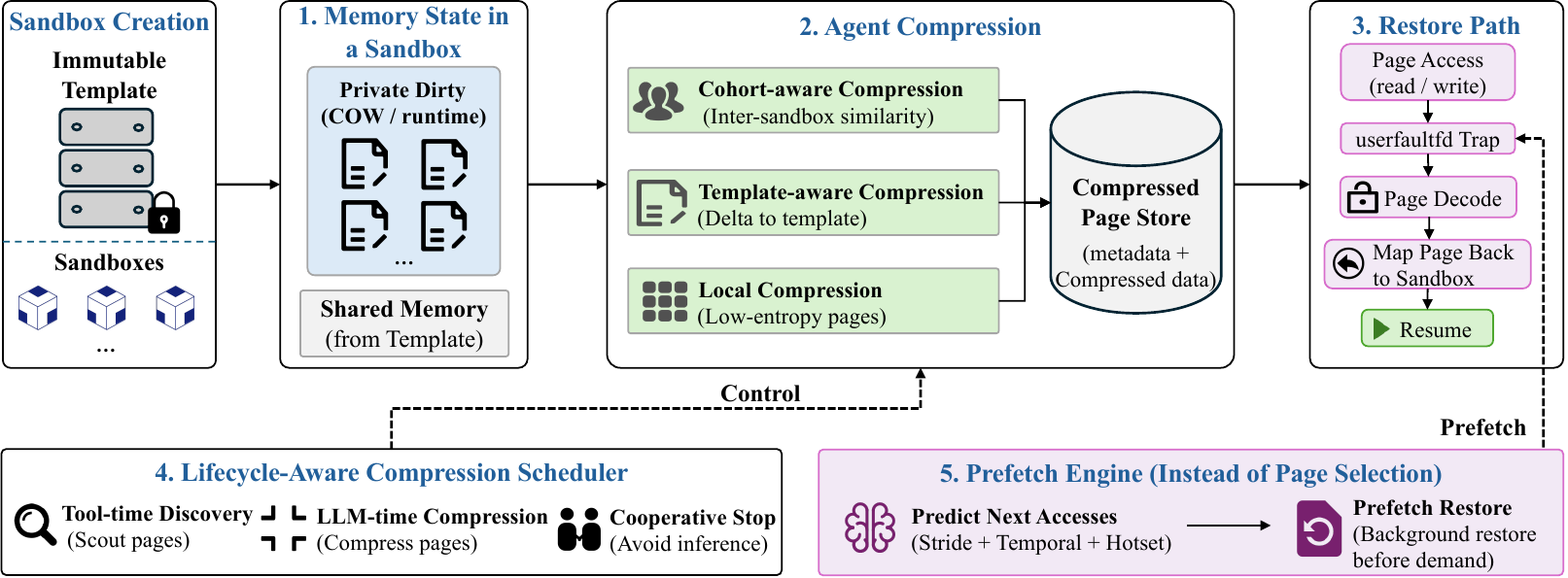}
\caption{AgentZip overview. AgentZip compresses sandbox-private pages using a portfolio of codecs, schedules expensive compression during LLM waiting periods, and proactively restores predicted pages before demand.}
\label{fig:overview}
\end{figure*}

Figure~\ref{fig:overview} shows AgentZip's high-level workflow. It consists of three main components. First, the \emph{compression algorithm} exploits sandbox-aware redundancy using a portfolio of cohort-aware, template-aware, and local compression methods. Second, the \emph{compression scheduler} decides when compression should run. Third, the \emph{restore prefetcher} learns page restore patterns and asynchronously restores likely-needed compressed pages before they trigger future demand page faults.

\subsection{Sandbox Model}
\label{sec}

AgentZip works based on a copy-on-write template model for AI-agent sandboxes~\cite{zeroboot}. In this model, sandboxes are created by forking or cloning from an immutable template that contains the base operating system state, runtime libraries, package environments, and pre-installed dependencies. Initially, sandbox pages can be shared with the template. When the guest modifies a shared page, copy-on-write creates a private sandbox-owned page. This model is common in agentic sandbox deployments because it enables fast sandbox creation and reduces the cost of storing identical initial states.

\subsection{Sandbox-Aware Compression Algorithm}
\label{sec}

Conventional memory compression systems typically compress pages in isolation, ignoring the structural redundancy exposed by AI-agent sandboxes. AgentZip instead designs sandbox-aware compression algorithms that explicitly exploit both template-relative and cross-sandbox similarity.

\textbf{Cohort-aware dictionary compression.}
It captures inter-sandbox similarity by organizing related sandboxes into a \textit{cohort}. AgentZip samples runtime pages from sandboxes in the same cohort and uses them to construct a shared immutable dictionary that captures common byte patterns. AgentZip uses this dictionary to encode candidate pages.

\textbf{Template-aware compression.}
This method is designed to capture template-to-sandbox similarity. Since sandboxes are created from a copy-on-write template, many private pages are dirty but still close to their corresponding template pages. AgentZip encodes them as deltas from the immutable template page at the same virtual page index.

\textbf{Local lightweight compression.}
This handles pages whose redundancy mainly exists within the page itself rather than across cohort or template sandboxes. For example, pages with repeated bytes or simple low-entropy patterns can be encoded efficiently by a local codec.

\subsection{Restore Prefetch Instead of Page Selection}
\label{sec}

Traditional memory compression systems rely on page selection to control performance overhead. They try to identify cold pages and avoid compressing hot pages. This reduces the chance of future page faults, but it also limits memory savings. In AI-agent sandboxes, this policy is too conservative. Many warm pages may still be highly similar to the template or to pages from other sandboxes. If AgentZip only compresses pages that are predicted to be cold, it would leave much of the sandbox-specific redundancy unused.

AgentZip therefore replaces traditional hotness-based page selection with prefetch-guided restore. The key idea is to decouple two decisions. The compression path asks whether a page has a profitable compressed representation. The restore path asks whether a compressed page is likely to be needed soon. This shifts the main performance-control mechanism from compression-time filtering to restore-time prediction, allowing AgentZip to compress more aggressively.

AgentZip draws inspiration from CPU prefetching~\cite{kim1997stride,ainsworth2024triangel,prophet,alecto,wu2021practical,mittal2016survey} and adapts them to predict compressed page restores in agent sandboxes. It uses multiple predictors to capture complementary restore patterns. A stride prefetcher captures regular sequential restore patterns. A temporal prefetcher captures page-to-page restore correlations, such as which page is likely to be restored after the current restored page. A request hotset prefetcher identifies pages that are frequently restored at the beginning of a tool-execution.

\subsection{Lifecycle-Aware Compression Timing}
\label{sec}

AgentZip targets reducing the average memory footprint of sandboxes and therefore treats compression as a lifecycle-aware service. The key idea is to separate \emph{tool-time discovery} from \emph{llm-time compression}. During tool execution, AgentZip runs a lightweight compressibility scout. The scout only observes and scores pages. It estimates potential compression savings and ranks high-value candidates. When the sandbox enters an idle period waiting for LLM thinking, AgentZip uses scout results to guide actual compression.

\section{AgentZip Design}

\subsection{Compression Infrastructure}
\label{sec:compression-infrastructure}

\begin{figure}[t]
\centering
\includegraphics[width=.99\linewidth]{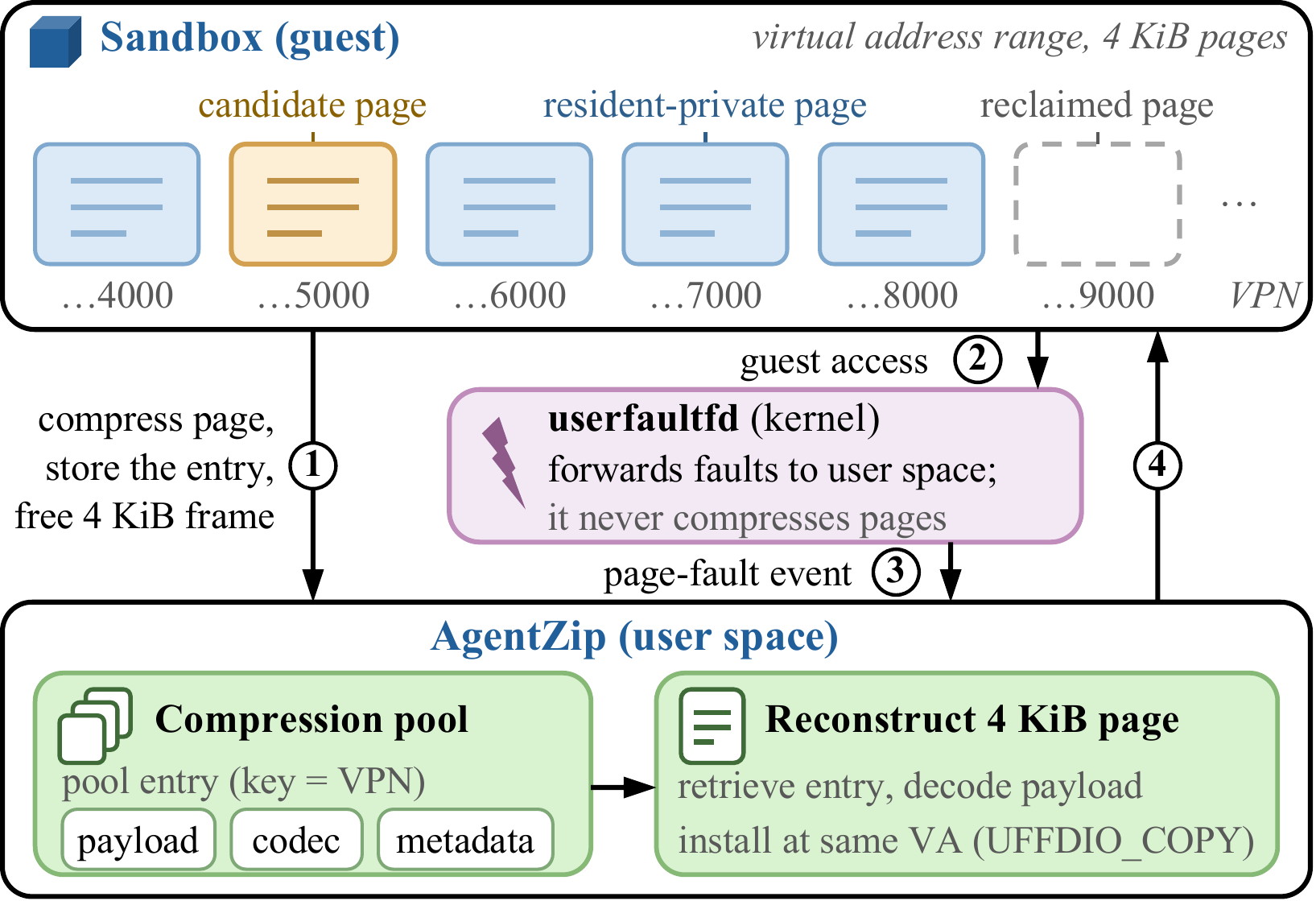}
\caption{AgentZip's compression infrastructure. Compressed pages are stored in a user-space pool. Later accesses to compressed pages are intercepted by \texttt{userfaultfd}.}
\vspace{-.2in}
\label{fig:compression_infra}
\end{figure}

As shown in Figure~\ref{fig:compression_infra}, AgentZip builds its compressed-memory infrastructure around a user-space compression pool and the Linux \texttt{userfaultfd} (UFFD)~\cite{linux-userfaultfd}. The compression pool stores compressed page contents, while UFFD allows AgentZip to detect and handle later accesses to these pages. UFFD can transfer page-fault handling from the kernel to AgentZip.

When AgentZip compresses a sandbox page, it stores a compressed representation in the user-space compression pool. Each pool entry contains the compressed payload, the compression algorithm used for encoding, and the associated metadata required to reconstruct the original page. After the compressed representation is safely stored, AgentZip releases the page's original 4\,KiB physical frame. We refer to such a page as a \emph{reclaimed page}. AgentZip registers the reclaimed page with UFFD. When the guest accesses it again, UFFD pauses the access and sends a page-fault event to AgentZip. AgentZip retrieves the compressed representation, reconstructs the original 4\,KiB content, and installs it back at the same address using \texttt{UFFDIO\_COPY}.

\subsection{Compression Algorithm}
\label{sec:design-compression}

AgentZip's compression has two stages. First, it provides a portfolio of three codecs that exploit complementary forms of redundancy in sandbox memory. Second, it evaluates the available codecs for each candidate page and selects the most space-efficient representation for final compression.

\subsubsection{Cohort Dictionary Compression}
\label{sec:design-dict}

Cohort dictionary compression captures redundancy across sibling sandboxes. AgentZip assigns each sandbox a cohort identifier as $\textit{cohort\_id}=\operatorname{Hash}(\textit{template\_id},\textit{request\_id})$. The \textit{template\_id} ensures that sandboxes in the same cohort share an identical base environment. The \textit{request\_id} groups sibling sandboxes created for the same agent task. Such sandboxes commonly execute similar commands, access similar files, and construct similar runtime states.

\textbf{Dictionary Training.} AgentZip samples resident-private pages from sandboxes in the same cohort and trains immutable Zstd dictionaries. The dictionary stores byte patterns that commonly appear in sandboxes within the cohort. Once a cohort has accumulated a sufficient number of samples, AgentZip initiates a new training round and partitions the sampled pages into a training set $\mathcal{T}$ and a validation set $\mathcal{V}$. AgentZip uses $\mathcal{T}$ to construct the candidate dictionary. It then evaluates each page on $\mathcal{V}$ by measuring the net memory saving, including both compressed sizes of the validation pages and storage cost of the dictionary itself.

AgentZip publishes a candidate dictionary $D^{*}$ only if it provides sufficient memory saving on $\mathcal{V}$ and fits within the global dictionary budget. Specifically, $D^{*}$ must satisfy $\mathrm{Saving}(D^{*},\mathcal{V}) \geq \theta_{\mathrm{dict}}$ and $M_{\mathrm{dict}} + M_{\mathrm{current}} \leq \theta_{\mathrm{total}}$, where $M_{\mathrm{dict}}$ is the size of $D^{*}$, $M_{\mathrm{current}}$ is the memory occupied by all currently published dictionaries, and $\theta_{\mathrm{total}}$ is the configured dictionary-memory budget. If both conditions hold, $D^{*}$ is published as the active dictionary. Otherwise, AgentZip retains the current active dictionary.

AgentZip always uses the most recently published dictionary as the active dictionary for subsequent compression, allowing it to capture byte patterns that better reflect the cohort's evolving runtime state. When a new dictionary becomes active, however, the previous dictionary cannot be removed immediately because pages compressed earlier may still depend on it for reconstruction. AgentZip therefore assigns each published dictionary a unique \textit{dict\_id}, and every dictionary-compressed page records the \textit{dict\_id} of the dictionary used during encoding. During restoration, AgentZip uses this identifier to retrieve the correct immutable dictionary. Each dictionary also maintains a reference count and is reclaimed only after no compressed page refers to it.

\textbf{Page Compression.}
When AgentZip processes an eligible 4\,KiB private page $P$, it retrieves the active dictionary $D$ of the cohort. AgentZip initializes a Zstd encoder with $D$ and encodes the page as $\operatorname{Zstd}_{D}(P)$.

\textbf{Page Reconstruction.}
For reconstructing a dictionary-compressed page, AgentZip reads the stored $\textit{dict\_id}$ and retrieves the exact dictionary used during compression. Let $B$ denote the stored compressed payload and let $D_{\mathrm{id}}$ denote the dictionary identified by the stored $\textit{dict\_id}$. AgentZip reconstructs the page as $\operatorname{Zstd}^{-1}_{D_{\mathrm{id}}}(B)$.

\subsubsection{Template-Delta Compression}
\label{sec:template_delta}

Template-delta compression exploits the copy-on-write structure of agent sandboxes. Each sandbox is derived from an immutable template that remains available throughout the lifetime of its descendants. Because the template is never modified after sandbox creation, template pages at each virtual page index provides a stable reference for both compression and reconstruction.

For a sandbox-private page at virtual page index $i$, let $P_i$ denote its current content and let $T_i$ denote the immutable template page at the same index. AgentZip stores the differences between $P_i$ and $T_i$ instead of storing the entire page.

\textbf{Page Compression.}
AgentZip divides each page into $N$ fixed-size blocks of $B$ bytes. We use $P_i^{(j)}$ and $T_i^{(j)}$ to denote the $j$-th blocks of the private page and template page, respectively. AgentZip constructs a change bitmap
\begin{equation*}
m_i^{(j)}
=
\begin{cases}
1, & P_i^{(j)} \neq T_i^{(j)},\\
0, & P_i^{(j)} = T_i^{(j)}.
\end{cases}
\label{eq:td-bitmap}
\end{equation*}

The encoded template delta is
\begin{equation*}
\Delta_i
=
\left(
M_i,
\left\{
P_i^{(j)}
\mid
m_i^{(j)} = 1
\right\}
\right),
\label{eq:td-encoding}
\end{equation*}
where $M_i$ is the change bitmap and $m_i^{(j)} = 1$ indicates that the $j$-th block differs from the corresponding template block.

\textbf{Page Reconstruction.}
To reconstruct page $P_i$, AgentZip first reads the immutable template page $T_i$. It then uses the change bitmap $M_i$ to determine the source of each block. An unchanged block is copied directly from the template, whereas a changed block is read from the stored delta payload. The complete page is reconstructed as \begin{equation*}
\widehat{P}_i
=
\mathop{\Vert}_{j=0}^{N-1}
\begin{cases}
T_i^{(j)}, & m_i^{(j)}=0,\\
\Delta_i^{(j)}, & m_i^{(j)}=1,
\end{cases}
\label{eq:td-page-reconstruction}
\end{equation*}
where $\Delta_i^{(j)}$ denotes the changed block stored in the delta payload and $\Vert$ denotes byte concatenation. 

\subsubsection{RLE Compression}
\label{sec:design-rle}

It captures repeated-byte redundancy within a single page. It is particularly effective for pages containing long regions with the same byte value, such as zero-like pages. 

\textbf{Page Compression.}
AgentZip scans a page $P$ from beginning to end and groups adjacent bytes with the same value. Suppose this process produces $m$ groups. For the $k$-th group, $b_k$ denotes the repeated byte value, and $\ell_k$ denotes the number of times that value appears consecutively. AgentZip stores each group using the pair $(b_k,\ell_k)$:
\begin{equation*}
\operatorname{RLE}(P)
=
\bigl(
(b_1,\ell_1),
(b_2,\ell_2),
\ldots,
(b_m,\ell_m)
\bigr).
\label{eq:rle-compression}
\end{equation*}

\textbf{Page Reconstruction.}
To reconstruct the page, AgentZip reads the encoded pairs in order. For each pair $(b_k,\ell_k)$, it writes byte value $b_k$ exactly $\ell_k$ times. It then concatenates all reconstructed groups to obtain the original page:
\begin{equation*}
\widehat{P}
=
\underbrace{b_1 b_1 \cdots b_1}_{\ell_1\ \mathrm{bytes}}
\Vert
\underbrace{b_2 b_2 \cdots b_2}_{\ell_2\ \mathrm{bytes}}
\Vert
\cdots
\Vert
\underbrace{b_m b_m \cdots b_m}_{\ell_m\ \mathrm{bytes}}
\label{eq:rle-reconstruction}
\end{equation*}

\subsubsection{Codec Selection and Admission}
\label{sec:design-codec-selection}

For each candidate page, AgentZip evaluates all available compression codecs. It computes the complete storage cost of each representation, including both the compressed payload and its metadata. AgentZip then selects the codec that produces the smallest representation for final compression:
\begin{equation*}
C^{*}(P)
=
\min
\left\{
C_{\mathrm{dict}}(P),
C_{\mathrm{TD}}(P),
C_{\mathrm{RLE}}(P)
\right\},
\label{eq:codec-selection}
\end{equation*}

\subsection{Restore Prefetch Algorithm}
\label{sec:design-prefetch}

Aggressive compression increases the likelihood that a reclaimed page will be accessed again, potentially triggering a page fault and increasing execution latency. To mitigate this cost, AgentZip learns page access patterns from previous restore events and proactively prefetches likely-needed compressed pages before they are accessed by the guest.

\subsubsection{Prefetch Input Stream}
\label{sec:prefetch-input}

AgentZip learns from demand page restores. A demand restore occurs when the guest accesses a page that is not in the physical memory. Let $p_t$ denote the page index restored by the $t$-th demand restore. The sequence $p_1, p_2, \ldots, p_t$ forms the input training stream.

\subsubsection{Prefetch Algorithms}
\label{sec:prefetch-algorithms}

AgentZip maintains three complementary prefetchers that capture different page restore patterns observed during sandbox execution.

\textbf{Stride prefetcher} captures regular restore sequences. After restoring two consecutive pages $p_{t-1}$ and $p_t$, AgentZip computes their page-index difference $s_t = p_t - p_{t-1}$. If the same stride is repeatedly observed, AgentZip predicts $p_t+s_t$ as a likely next page. 

\textbf{Temporal prefetchers} captures page-to-page restore correlations. When demand restore $p_t$ follows $p_{t-1}$, AgentZip records the transition $p_{t-1} \rightarrow p_t$. For each source page, AgentZip maintains a set of successor pages together with their observation counts and recency. AgentZip maintains two temporal predictors that learn these transitions at different scopes: a local temporal predictor for each sandbox and a cohort temporal predictor shared by sibling sandboxes.

\textit{Local temporal predictor.} It records transitions observed within an individual sandbox. When page $p_t$ is restored, AgentZip looks up the successors previously observed after $p_t$ in the same sandbox. This prefetcher captures restore sequences that repeat within a trajectory and adapts to the sandbox's own execution history. 

\textit{Cohort temporal predictor.} It shares transition history across sandboxes with same \texttt{cohort\_id}. Each demand transition observed in one sandbox updates a cohort-level successor set. This prefetcher is useful when sibling sandboxes execute related tool sequences, because a restore pattern can be learned in one sandbox before it is encountered in another.

\textbf{Tool-call hotset prefetcher.}
Sandboxes within the same cohort often access similar pages at nearly tool-call positions in their agent trajectories. AgentZip maintains a tool-call-specific restore history indexed by $(\textit{cohort\_id},\textit{tool\_ordinal})$. For cohort $c$ and tool-call position $a$, it records the number of restores of each page $p$: $H_{c,a}(p) = \#\{\text{restores of } p \text{ at } (c,a)\}$. Before another sandbox begins its $a$-th tool call, AgentZip prefetches top-$K$ pages according to their restore frequency.

\subsubsection{Prefetch Request Generation.}

AgentZip generates a prefetch request only when the page is marked as compressed. A valid candidate is inserted into a bounded asynchronous prefetch queue. For each queue entry, AgentZip retrieves the corresponding compressed metadata, and decodes the page. After reconstructing it, AgentZip uses UFFD to copy the page back into its physical memory and changes its state to resident-private.

\subsection{Lifecycle-Aware Compression Scheduler}
\label{sec:design-scheduler}

\begin{figure}[t]
\centering
\includegraphics[width=.99\linewidth]{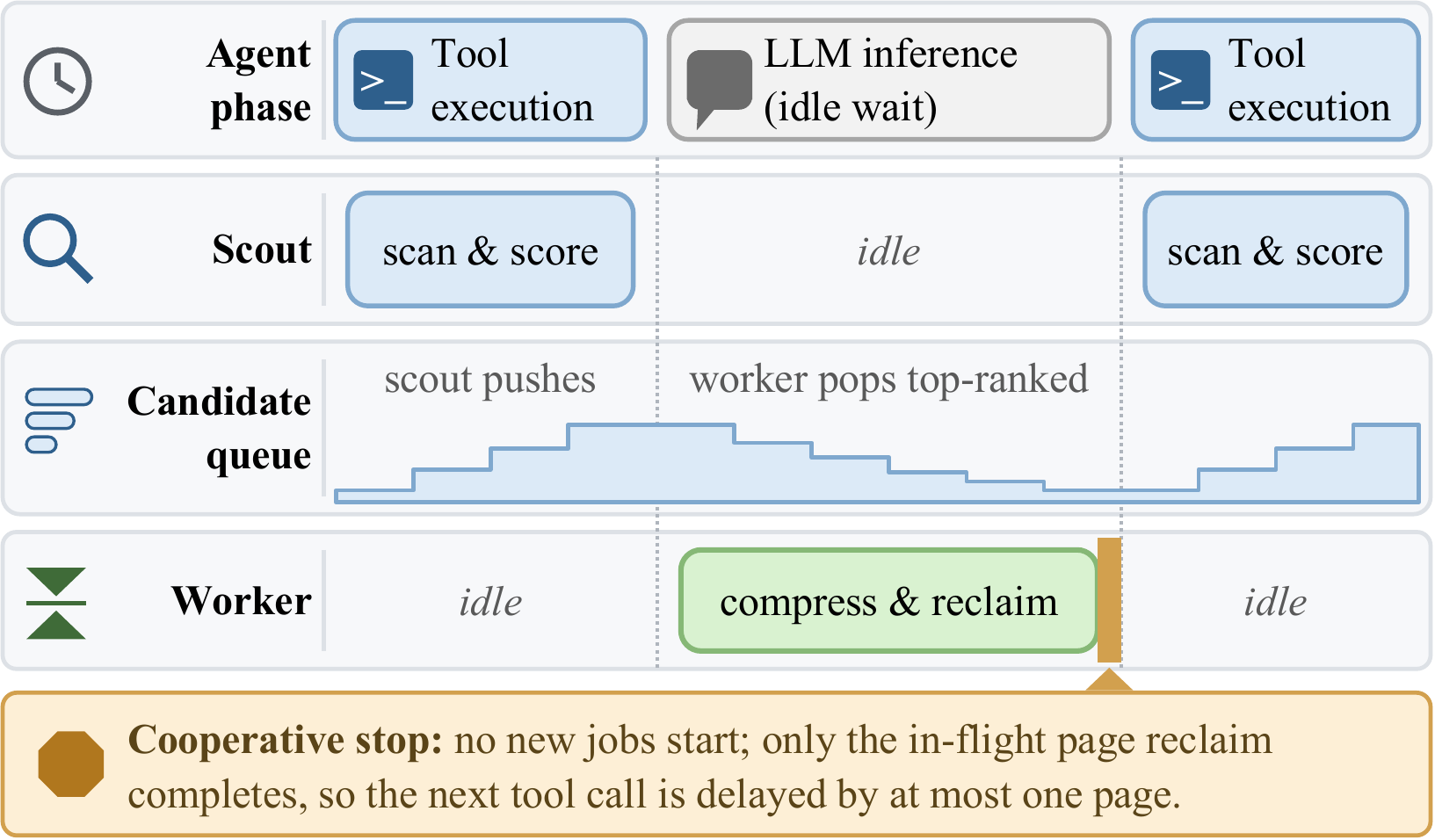}
\caption{AgentZip lifecycle-aware scheduler. A lightweight scout identifies compression candidates during tool execution, while expensive compression during LLM waiting.}
\vspace{-.2in}
\label{fig:lifecycle_scheduler}
\end{figure}

Agent workloads alternate between two distinct phases: tool execution, during which the sandbox actively consumes CPU, and LLM inference, during which the sandbox waits for the next model response. As shown in Figure~\ref{fig:lifecycle_scheduler}, AgentZip exploits this lifecycle by performing only lightweight page-candidate discovery during tool execution and deferring expensive page compression to subsequent LLM inference.

\textbf{Tool-time discovery.}
While a tool is running, AgentZip uses a lightweight scout to asynchronously evaluate the compression benefits of resident-private pages. For each page, the scout estimates how many bytes could be saved by compression. Pages with high estimated savings are placed in a candidate queue and ranked by their expected benefit.

\textbf{LLM-time compression.} After a tool call completes, AgentZip performs compression while the sandbox waits for the next LLM response. It processes the pages identified by the scout in descending order of estimated memory saving. For each candidate, AgentZip applies the most space-efficient codec, stores the compressed page in the compression pool, and then reclaims the page's resident physical frame.

\textbf{Cooperative stop.} When a new tool call arrives, AgentZip sends a cancellation request to the compression worker. Upon receiving the request, the worker stops launching new compression operations. Any candidate that has not yet modified the page mapping is abandoned, and any temporary compressed representation is released. An ongoing page-mapping update, however, cannot be safely interrupted. If cancellation arrives while AgentZip is reclaiming a page, the worker completes that single-page update before stopping. AgentZip serializes page-mapping updates so that guest execution never overlaps with an incomplete reclamation.

\subsection{E2B API Compatibility}
\label{sec:design-e2b}

E2B provides a widely adopted sandbox interface for AI-agent workloads~\cite{e2b,openai-sandbox-clients}, exposing persistent Linux environments in which agents can execute commands, manipulate files, and preserve workspace state across tool calls. AgentZip provides an E2B-compatible frontend so that existing E2B-based agents can use AgentZip without modifying their prompts, tools, or orchestration logic.

The E2B interface includes a control-plane API for sandbox lifecycle management and a per-sandbox \textit{envd} API for command execution. AgentZip supports the subset required by our workloads through a host-side compatibility gateway. Existing E2B clients redirect their API endpoints to this gateway while continuing to use the original E2B SDK.

\textbf{Control-plane compatibility.} The gateway translates \texttt{Sandbox.create} requests into AgentZip sandbox creation. The E2B \textit{templateID} identifies the immutable template used to instantiate the persistent copy-on-write sandbox, while request metadata provides the \textit{request\_id}. AgentZip derives the corresponding \textit{cohort\_id} from the template and request identifiers for cohort dictionary compression and cohort-level prefetching. The gateway maintains a mapping between each E2B \textit{sandboxID} and its AgentZip sandbox for subsequent lifecycle operations. AgentZip further namespaces sandbox, dictionary, and predictor state by tenant to prevent sharing across mutually untrusted users.

\textbf{Command compatibility.} The gateway terminates E2B's \textit{envd} command interface and translates each \texttt{commands.run} request into an AgentZip guest-execution request. Command arguments, working directory, environment variables, user, and timeout are forwarded to the corresponding persistent sandbox, while the resulting standard output, error, and exit status are translated back into the format expected by E2B SDK. This translation preserves E2B programming interface while reusing AgentZip's native sandbox execution path.

\section{Experimental Methodology}
\label{sec:methodology}

\subsection{Experimental Setup}
\label{sec:experimental-setup}

We implement AgentZip on top of Zeroboot~\cite{zeroboot}, a KVM-based sandbox runtime that creates sandboxes from Firecracker memory snapshots using CoW cloning. We conduct experiments on a server equipped with Intel Xeon Platinum 8480C processors, running Linux 5.15.0. All sandboxes belonging to the same workload use the same snapshot as their template and initially share its memory through copy-on-write mappings. Pages modified during execution become sandbox-owned private pages.

\subsection{Agent Workloads}
\label{sec:agent-workloads}

We construct the workloads from ten Python repositories in R2E-Gym~\cite{jain2025r2e}: \texttt{aiohttp}, \texttt{coverage.py}, \texttt{DataLad}, \texttt{NumPy}, \texttt{Orange3}, \texttt{pandas}, \texttt{Pillow}, \texttt{Pyramid}, \texttt{Scrapy}, and \texttt{Tornado}. To evaluate AgentZip under both LLM training and LLM inference settings, we use R2E-Gym to construct two representative sandbox workload patterns: parallel rollout for training and generate-and-filter (GAF) for inference.


\textbf{Parallel rollout.} We emulate this scenario by generating 16 independent trajectories for each R2E task using DeepSeek-V4~\cite{xu2026deepseek}. All trajectories start from the same issue prompt and repository state. We use different LLM temperatures to obtain diverse tool-use behaviors. The 16 trajectory sandboxes execute concurrently and share a cohort identifier.


\textbf{Generate-and-filter.} We emulate this scenario by generating four candidate trajectories for each selected R2E task. All candidates receive the same issue prompt, but use different role prompts (e.g., minimal patch, failure reproduction, test-guided repair, and root-cause diagnosis), and decoding temperatures. These settings encourage the candidates to follow different execution paths. The four trajectories are replayed concurrently in separate persistent sandboxes and share one cohort identifier. 

\subsection{Trace Replay}

\label{sec:trace-replay}

We evaluate AgentZip by replaying traces collected from representative agent workloads. Each trace records the sequence of tool actions together with the LLM inference latency preceding each tool call. To ensure fair comparison, each trace is generated once and replayed identically across all evaluated configurations. Before issuing a tool action, the replay waits for the recorded \texttt{think\_time\_ms\_before} to reproduce the original LLM inference interval. During this period, the sandbox is marked as LLM-idle, allowing AgentZip to perform background compression.

\subsection{Memory Measurement}
\label{sec:memory-measurement}

Our primary memory metric is \emph{sandbox-owned physical memory}. At time $t$, we define it as
\begin{equation*}
M(t)=B_{\mathrm{private}}(t)+B_{\mathrm{pool}}(t),
\label{eq:sandbox-owned-memory}
\end{equation*}
where $B_{\mathrm{private}}(t)$ denotes the physical memory occupied by sandbox-private pages, and $B_{\mathrm{pool}}(t)$ denotes the physical memory occupied by the user-space compression pool, including compressed page contents and associated metadata.

We record $M(t)$ once per second throughout each workload. For every workload $w$, we derive $M_w^{\mathrm{avg}}$: the time-averaged sandbox-owned memory. The memory saving of AgentZip relative to its paired no-compression execution is 
\begin{equation*}
S_w^{x}
=
1-
\frac{M_{\mathrm{AgentZip},w}^{x}}
     {M_{\mathrm{NoComp},w}^{x}},
\label{eq:paired-saving}
\end{equation*}

\subsection{Latency Measurement}
\label{sec:latency-measurement}

For each configuration, \emph{wall time} is measured from the start of concurrent trajectory replay until the last sandbox completes. For workload $w$, we define AgentZip slowdown as
\begin{equation*}
L_w
=
\frac{T_{\mathrm{AgentZip},w}}
     {T_{\mathrm{NoComp},w}},
\label{eq:workload-slowdown}
\end{equation*}

\begin{figure*}[t]
  \centering
  \includegraphics[width=\textwidth]
  {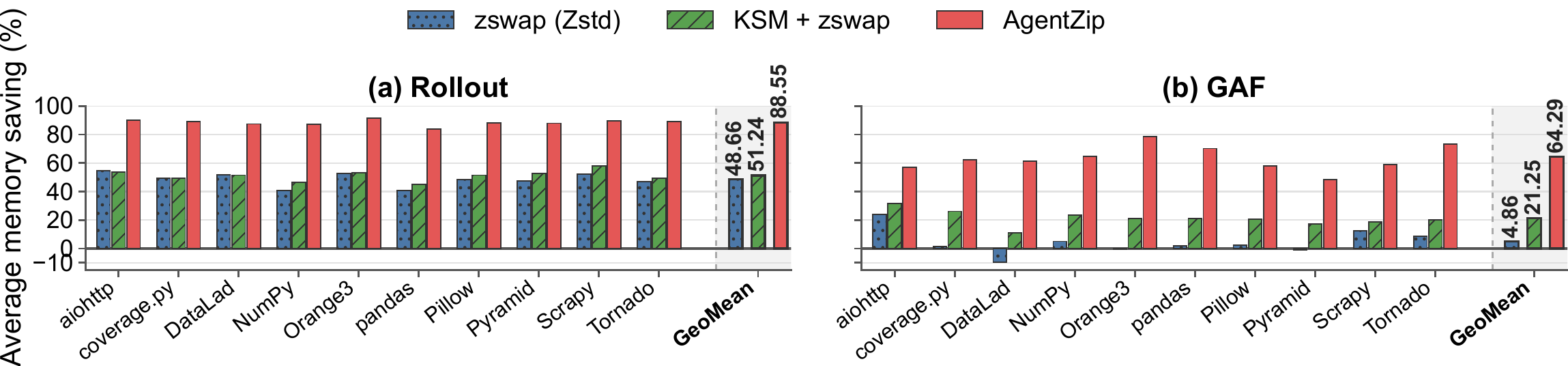}
  \vspace{-.25in}
  \caption{Average sandbox-owned memory saving relative to the paired no-compression baseline across Rollout and GAF.}
  \vspace{-.15in}
  \label{fig:overall-saving}
\end{figure*}

\begin{figure*}[t]
  \centering
  \includegraphics[width=\textwidth]{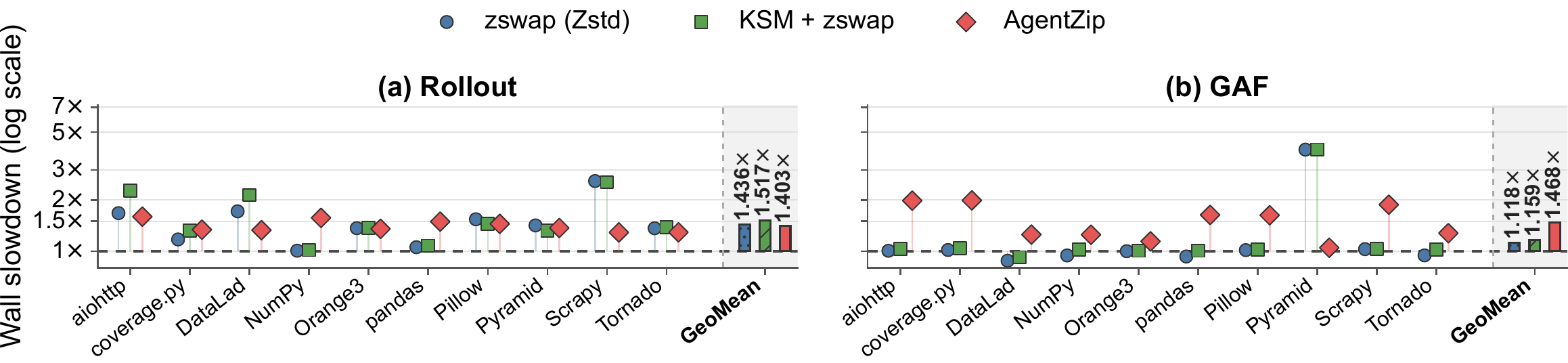}
  \vspace{-.25in}
  \caption{End-to-end wall-time slowdown relative to the paired no-compression baseline across Rollout and GAF.}
  \vspace{-.15in}
  \label{fig:overall-wall}
\end{figure*}

\section{Evaluation}
\label{sec:evaluation}

\subsection{Overall Results}
\label{sec:eval-overall}

Figure~\ref{fig:overall-saving} shows that AgentZip substantially reduces the sandbox memory footprint in both LLM training and inference workloads. On Rollout, AgentZip reduces average sandbox-owned memory by 88.55\%, compared with 48.66\% for zswap and 51.24\% for KSM+zswap. On GAF, AgentZip achieves a 64.29\% reduction, whereas zswap and KSM+zswap reduce memory by only 4.86\% and 21.25\%, respectively.

Figure~\ref{fig:overall-wall} reports the corresponding end-to-end latency. On Rollout, AgentZip runs at 1.403$\times$ the no compression (\textsc{NoComp}) wall time, lower than both zswap at 1.436$\times$ and KSM+zswap at 1.517$\times$. On GAF, AgentZip runs at 1.468$\times$ the \textsc{NoComp} wall time, compared with 1.118$\times$ for zswap and 1.159$\times$ for KSM+zswap. On Rollout, AgentZip therefore achieves both the largest memory saving and the lowest slowdown among the evaluated compression schemes, demonstrating the best overall memory-latency trade-off. Although AgentZip incurs moderately higher latency on GAF, it delivers substantially greater memory savings.

\textbf{Zswap: reactive compression timing.} The primary limitation of zswap lies in its reactive trigger: it compresses pages only under memory pressure. This late trigger is poorly matched to short-lived agent sandboxes, where compressible pages may remain resident for most or all of a request and continue to contribute to the average memory footprint. AgentZip instead proactively identifies compressible pages during tool execution and reclaims them during LLM waiting intervals, decoupling compression from memory pressure.

\textbf{KSM+zswap: sandbox-oblivious redundancy detection.} Most exact sharing among sibling sandboxes is already captured by template-backed copy-on-write mappings at sandbox creation. Once a write creates a private page, KSM can merge it only if its entire content is byte-identical to another page. Even small differences prevent sharing despite substantial template-relative or cross-sandbox similarity. AgentZip targets this residual redundancy with template-delta compression for pages that remain close to their template versions and cohort dictionary compression for shared byte patterns across sibling sandboxes.

\subsection{Effectiveness of the Codec Portfolio}
\label{sec:eval-codecs}

\begin{figure}[t]
  \centering
  \includegraphics[width=\linewidth]{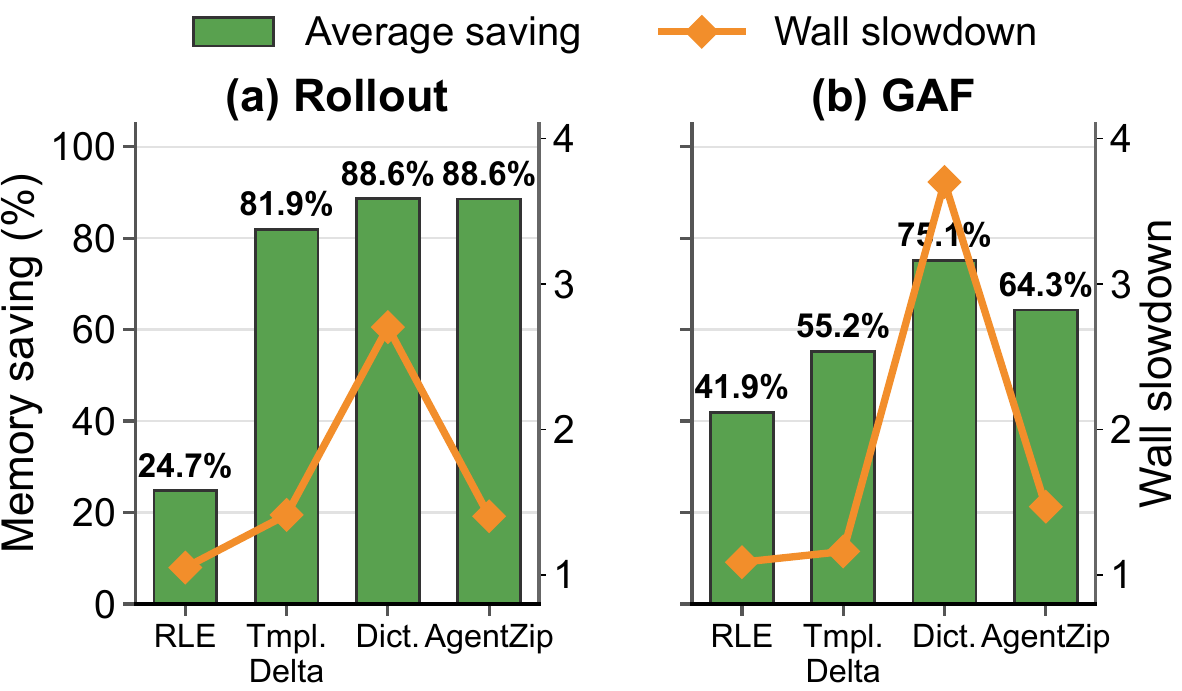}
  \vspace{-.2in}
  \caption{Effectiveness of AgentZip's codec portfolio.}
  \vspace{-.2in}
  \label{fig:codec-ablation}
\end{figure}

Figure~\ref{fig:codec-ablation} compares the memory savings achieved by each compression algorithm in isolation. RLE reduces average sandbox-owned memory by 24.74\% on Rollout and 41.93\% on GAF, indicating that repeated-byte and zero-like pages provide a meaningful but limited compression opportunity. Template-delta compression achieves substantially higher savings of 81.95\% on Rollout and 55.24\% on GAF, showing that many private copy-on-write pages remain similar to their corresponding immutable template pages. Cohort dictionary compression reduces average memory by 88.63\% on Rollout and 75.14\% on GAF, which confirms that substantial byte-level redundancy persists across sibling sandboxes even as their execution states diverge.

The key insight is that combining complementary compression methods can substantially reduce compression cost while preserving most of the achievable memory saving. AgentZip uses lightweight RLE and template-delta compression whenever possible, while reserving cohort dictionary compression for pages that require cross-sandbox context. On Rollout, this reduces the number of dictionary-encoded pages by about 60\%, while achieving nearly the same memory saving as dictionary-only compression (88.55\% versus 88.63\%) and reducing wall-time slowdown from 2.703$\times$ to 1.403$\times$. This demonstrates that the codec portfolio preserves broad compression coverage while substantially reducing the cost of dictionary-based compression. 

\subsection{Lifecycle-Aware Compression Timing}
\label{sec:eval-scheduler}

\begin{figure}[t]
  \centering
  \includegraphics[width=\linewidth]{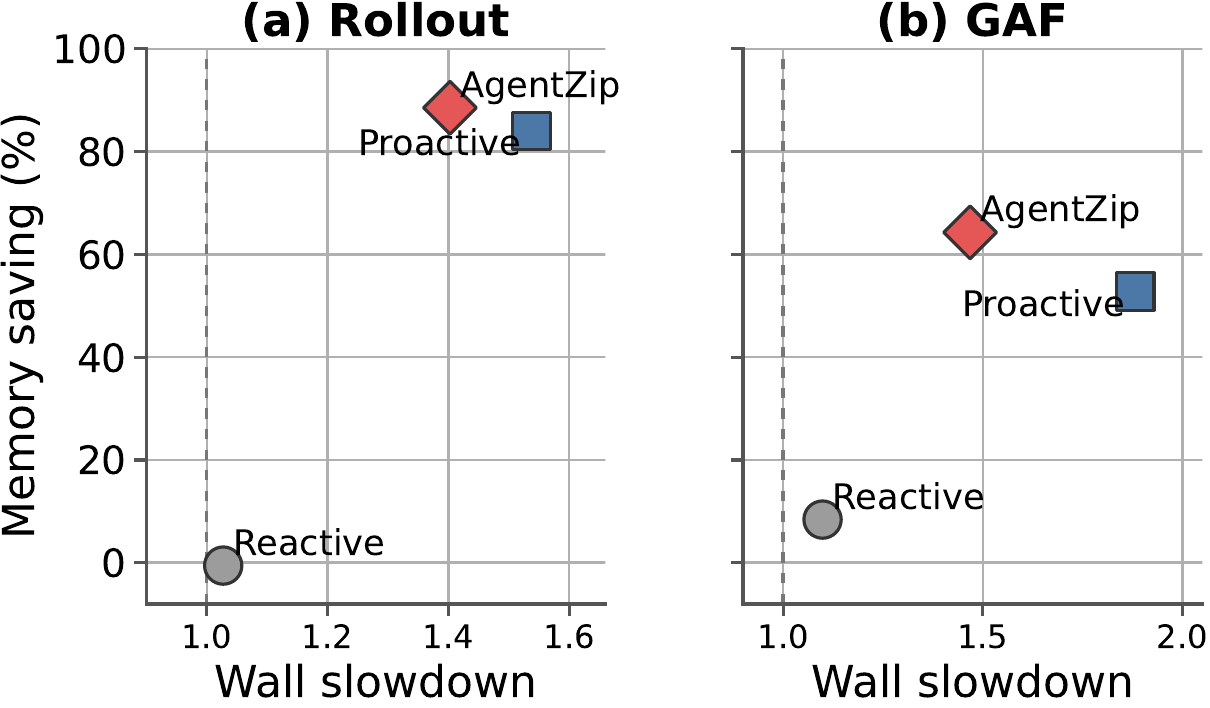}
  \vspace{-.1in}
  \caption{Memory--latency trade-off of compression timing under the same
  compression algorithm portfolio.}
  \vspace{-.15in}
  \label{fig:scheduler-tradeoff}
\end{figure}

Figure~\ref{fig:scheduler-tradeoff} compares AgentZip with reactive compression and conventional proactive compression strategies discussed in Section~\ref{subsec:back_memcomp}. On Rollout, reactive strategy achieves essentially no average memory saving ($-0.56$\%) at 1.027$\times$ wall time, while the proactive strategy reduces average memory by 84.04\% at 1.538$\times$ wall time. AgentZip achieves the highest saving of 88.55\% while reducing wall time to 1.403$\times$. On GAF, the reactive strategy reduces average memory by only 11.73\%. The generic strategy achieves 52.78\% memory saving at 1.882$\times$ wall time, whereas AgentZip increases memory saving to 64.29\% while reducing wall time to 1.468$\times$.

These results demonstrate the benefit of aligning compression with the agent lifecycle. The reactive strategy compression waits for memory pressure and therefore misses most opportunities to reduce the time-averaged footprint of short-lived sandboxes. The conventional proactive strategy exposes more compression opportunities, but performs expensive scanning, encoding, and page reclamation without distinguishing foreground tool execution from LLM waiting periods. AgentZip instead performs lightweight discovery during tool execution and moves expensive compression and mapping updates into LLM waiting windows. As a result, AgentZip achieves a better memory--latency operating point than proactive strategy on both workloads.

\subsection{Effectiveness of Restore Prefetching}
\label{sec:eval-prefetch}

\begin{figure}[t]
  \centering
  \includegraphics[width=\linewidth]{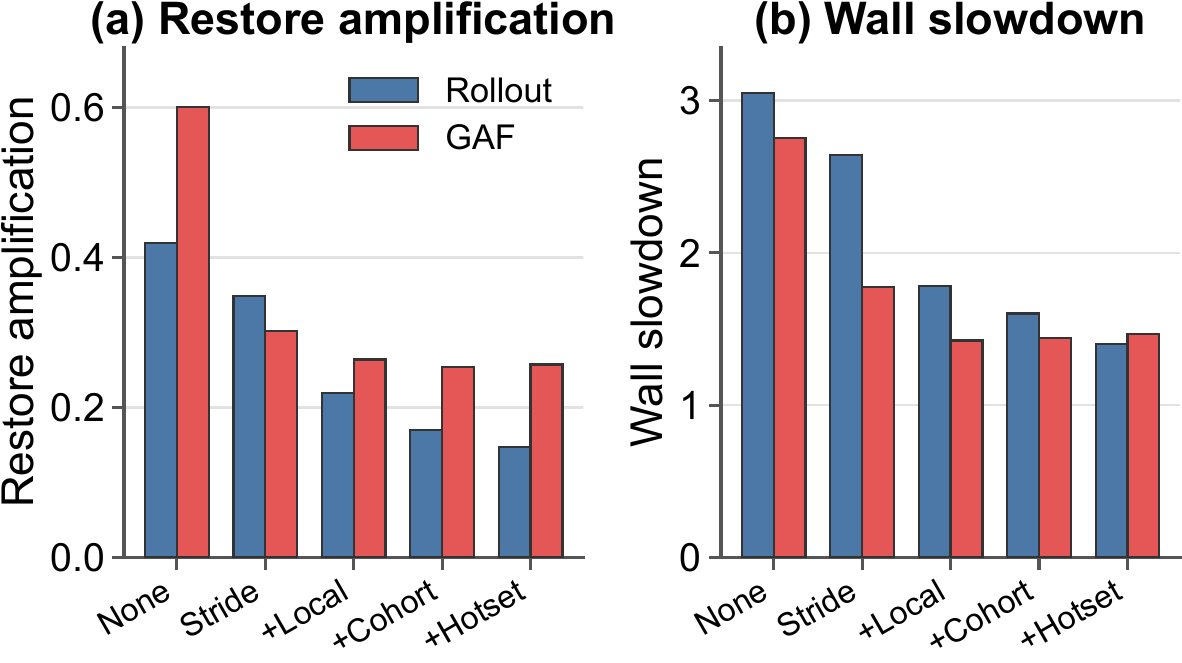}
  \vspace{-.1in}
  \caption{Cumulative prefetch ablation. (a) Demand restores amplification normalized by the number of reclaimed pages. (b) End-to-end slowdown.}
  \vspace{-.1in}
  \label{fig:prefetch-effectiveness}
\end{figure}

\begin{figure}[t]
  \centering
  \includegraphics[width=\linewidth]{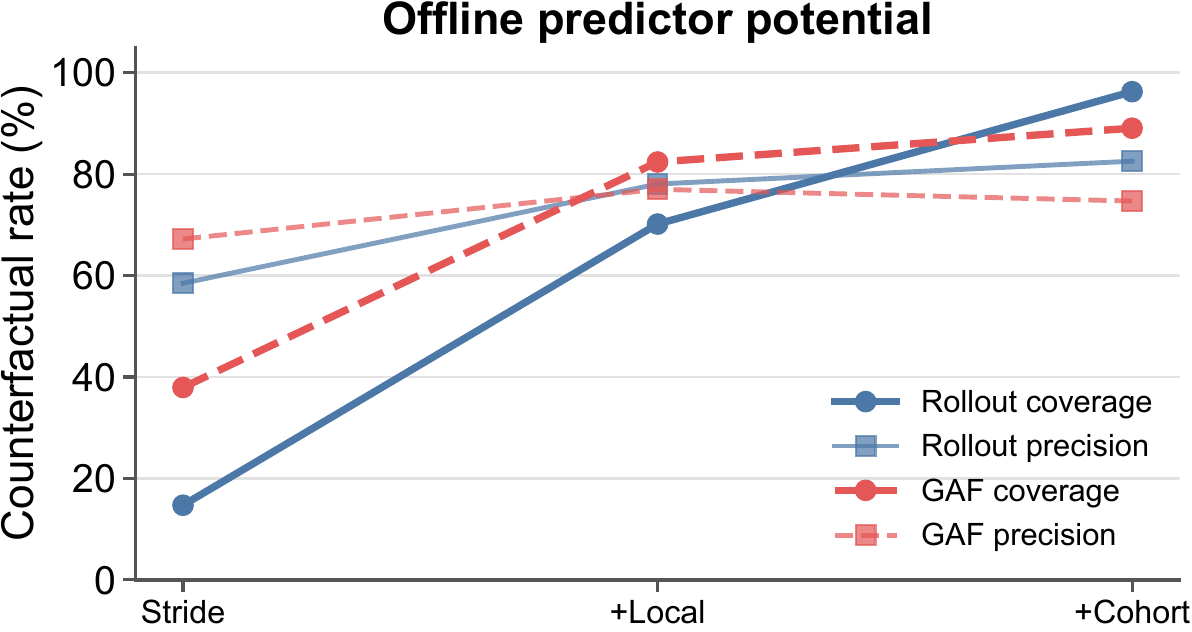}
  \caption{Counterfactual coverage and precision.}
  \vspace{-.2in}
  \label{fig:prefetch-potential}
\end{figure}

We first quantify the performance cost of restoring compressed pages using restore amplification, defined as

\begin{equation*}
A_{\mathrm{restore}}=\frac{N_{\mathrm{demand\ restores}}}{N_{\mathrm{reclaimed\ pages}}},
\end{equation*}

This metric captures how frequently reclaimed pages are subsequently accessed through blocking demand restores: a higher value indicates that more compressed pages are later brought back onto the guest's critical path.

Figure~\ref{fig:prefetch-effectiveness} reports the ablation study results of restore prefetch. Without prefetching, AgentZip achieves 94.22\% average memory saving on Rollout and 75.96\% on GAF, but restore amplification reaches 0.419 and 0.600, resulting in wall-time slowdowns of 3.052$\times$ and 2.755$\times$, respectively. On Rollout, adding stride, local temporal, cohort temporal, and tool-hotset prefetcher progressively reduces restore amplification from 0.419 to 0.349, 0.219, 0.170, and 0.147. The corresponding wall-time slowdown decreases from 3.052$\times$ to 2.645$\times$, 1.784$\times$, 1.602$\times$, and finally 1.403$\times$. On GAF, stride and local temporal prefetcher reduce restore amplification from 0.600 to 0.302 and 0.264, while reducing wall-time slowdown from 2.755$\times$ to 1.776$\times$ and 1.424$\times$. Cohort temporal prefetcher further reduces restore amplification to 0.254, and the complete AgentZip prefetcher reaches 0.257 at 1.468$\times$ wall time.

Figure~\ref{fig:prefetch-potential} further evaluates the prediction opportunity using demand restore traces. On Rollout, stride prefetcher covers 14.72\% of future demand restores with 58.49\% precision. Adding local temporal prefetcher increases coverage to 70.16\% with 78.05\% precision, while cohort temporal prefetcher further increases coverage to 96.27\% with 82.56\% precision. On GAF, coverage increases from 37.94\% with stride prefetcher to 82.39\% with local temporal prefetcher and 89.06\% with cohort temporal prefetcher, with the complete temporal predictor achieving 74.70\% precision.

These results demonstrate the key role of AgentZip's restore prefetching. Aggressive compression exposes more memory-saving opportunities by compressing warm pages that conventional policies would normally retain. However, these pages are likely to be accessed again, triggering blocking UFFD restores that stall guest execution. By predicting which compressed warm pages will be needed again and restoring them before demand, AgentZip can compress warm pages aggressively without treating their likely reuse as a reason to leave them uncompressed. AgentZip makes these predictions using complementary prefetchers that exploit different access patterns in agent workloads.

\subsection{Similarity and Dictionary Sensitivity}
\label{sec:eval-similarity}

\begin{figure}[t]
  \centering
  \includegraphics[width=\linewidth]{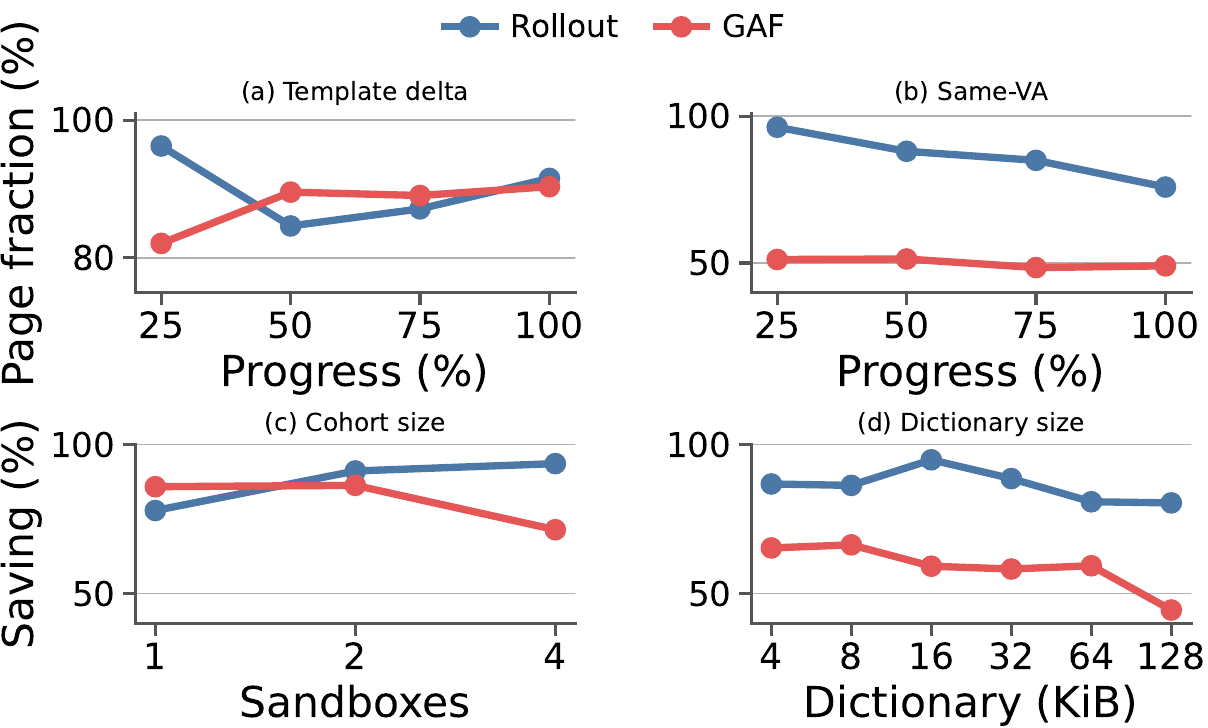}
  \vspace{-.2in}
  \caption{Similarity and dictionary sensitivity study.}
  \vspace{-.2in}
  \label{fig:similarity-sensitivity}
\end{figure}

AgentZip's compression design is motivated by two sources of memory
similarity: \emph{template-relative similarity} and \emph{cross-sandbox similarity}. This experiment answers two questions: whether these similarities remain present as agent executions progress, and how the cohort size and dictionary size affect AgentZip's ability to convert them into memory savings. Figures~\ref{fig:similarity-sensitivity}(a) and~(b) validate template-relative and cross-sandbox similarity, respectively, while Figure~\ref{fig:similarity-sensitivity}(c) and~(d) evaluate the configuration of the
cohort dictionary used to exploit cross-sandbox redundancy

Figures~\ref{fig:similarity-sensitivity}(a) and (b) show that substantial redundancy persists throughout agent execution. On Rollout, 96.26\% of pages contain the same content as the sibling sandboxes at the corresponding virtual address at 25\% of execution, and this fraction remains 75.87\% at completion. This shows that sibling trajectories retain substantial memory similarity even after their executions diverge. GAF shows lower cross-sandbox agreement, with about 48--51\% of pages matching the majority content at the corresponding virtual address, reflecting the greater diversity of independently generated candidates. Nevertheless, template-relative similarity remains high in both workloads: the fraction of pages profitably compressible with template delta ranges from 84.64\% to 96.25\% on Rollout and increases from 82.10\% to 90.34\% on GAF. These results validate AgentZip's use of complementary compression mechanisms: cohort dictionaries exploit redundancy shared across sibling sandboxes, while template delta continues to capture private pages even when their exact contents have diverged.

Figures~\ref{fig:similarity-sensitivity}(c) and (d) show that cross-sandbox
dictionary compression benefits from choosing both cohort size and dictionary
capacity carefully. On Rollout, dictionary-only memory saving increases from 77.83\% with one sandbox to 91.08\% with two and 93.55\% with four, indicating
that additional sibling sandboxes provide useful shared patterns. GAF shows a
different trend: saving increases only slightly from 85.81\% with one candidate to 86.26\% with two, but drops to 71.39\% with four, suggesting that more diverse candidates introduce patterns that are less useful for the pages being compressed. Dictionary capacity shows a similar trade-off. Rollout achieves its highest saving of 94.83\% with a 16-KiB dictionary, while GAF peaks at 66.31\% with 8~KiB; increasing the dictionary to 128~KiB reduces saving to 80.38\% and 44.46\%, respectively. Thus, adding more sandboxes or allocating more dictionary space does not necessarily improve compression. AgentZip addresses this trade-off by validating candidate dictionaries on held-out samples and accounting for dictionary storage before publication, rather than simply using the largest available cohort or dictionary.

\subsection{Memory Overheads}
\label{sec:eval-overheads}

\begin{figure}[t]
  \centering
  \includegraphics[width=\linewidth]{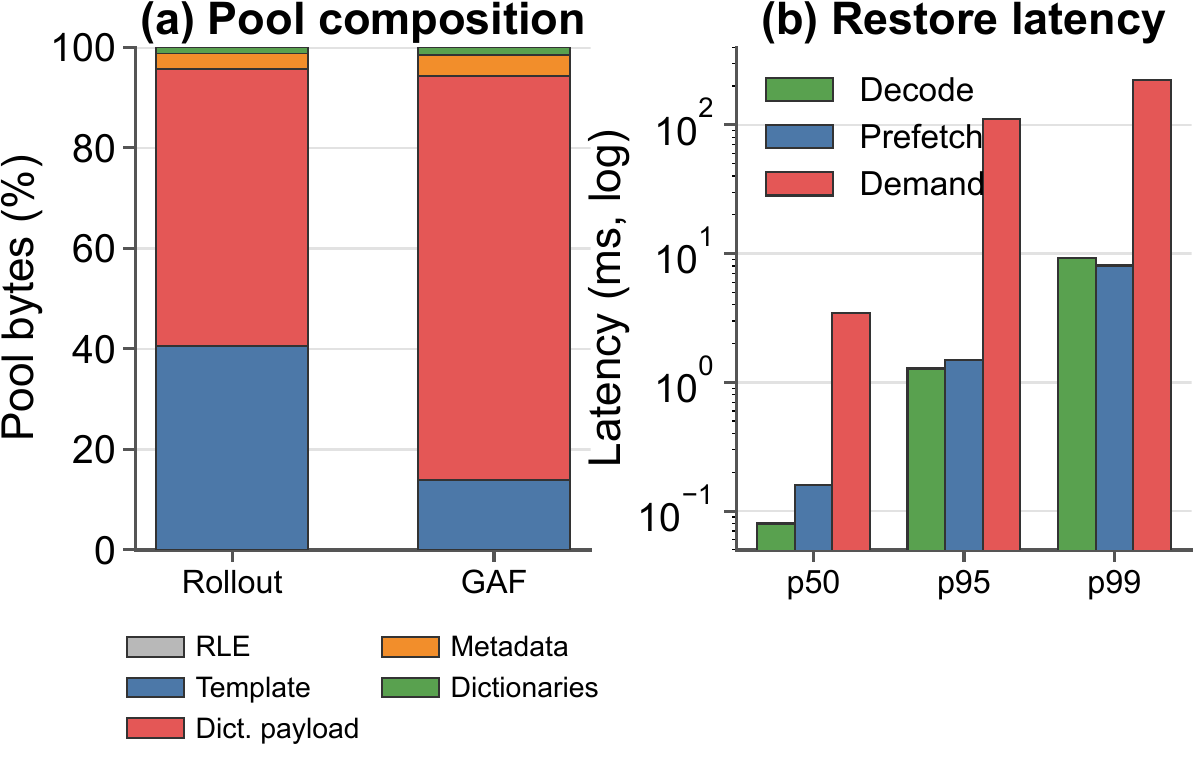}
  \vspace{-.2in}
  \caption{AgentZip compression and restore overheads. (a) Breakdown of compression-pool memory usage. (b) Latency of codec decoding, asynchronous prefetch restoration, and blocking demand restoration.}
  \vspace{-.2in}
  \label{fig:overheads}
\end{figure}

Figure~\ref{fig:overheads}(a) reports the memory overhead of AgentZip's compression pool. On Rollout, template-delta and dictionary payloads account for 40.56\% and 55.14\% of pool bytes, while blob metadata and immutable dictionaries account for only 3.09\% and 1.22\%, respectively. On GAF, dictionary and template-delta payloads account for 80.45\% and 13.88\%, while metadata and immutable dictionaries account for 4.09\% and 1.55\%. Overall, auxiliary metadata and dictionary storage constitute only 4.31\% of the pool on Rollout and 5.64\% on GAF. The complete compression pool, including all payload and metadata overheads, further accounts for 28.9\% and 6.1\% of profiled sandbox-owned memory on Rollout and GAF.

Figure~\ref{fig:overheads}(b) reports the restore-path latency for compressed pages. Codec decoding takes 0.08~ms, 1.28~ms, and 9.21~ms at p50, p95, and p99, respectively, while asynchronous prefetch restoration takes 0.16~ms, 1.50~ms, and 8.09~ms. In contrast, demand restoration takes 3.47~ms at p50, 110.52~ms at p95, and 223.27~ms at p99.

These results show that AgentZip introduces little auxiliary memory overhead: more than 94\% of the compression pool stores useful compressed page contents rather than metadata or dictionaries. More importantly, codec reconstruction itself is inexpensive compared with a blocking demand restore, whose tail latency is over an order of magnitude larger. This distinction validates AgentZip's design choice to focus on restore prefetching rather than only optimizing compression and decompression latency: moving restoration off the demand-fault critical path provides substantially greater performance benefit while preserving the large memory savings enabled by aggressive compression.
\section{Conclusion}
\label{sec:conclusion}

High-fanout agent workloads make sandbox memory a first-order scaling bottleneck, but also expose redundancy and execution structure that conventional memory-management mechanisms fail to exploit. AgentZip exploits these properties by compressing redundancy across sibling sandboxes and against their shared template, aligning expensive compression with LLM waiting periods, and predicting future restores so that warm pages can be compressed aggressively without relying on conservative page selection. Together, these techniques show that agent sandboxes should not be treated as independent general-purpose workloads: their shared origins, correlated executions, and repeated idle intervals provide system-level opportunities to substantially increase deployment density while controlling the performance cost of aggressive memory compression.

\bibliographystyle{ACM-Reference-Format}
\bibliography{refs}

@misc{e2b,
  author       = {{E2B}},
  title        = {{E2B}: Secure Sandboxes for AI Agents},
  year         = {2026},
  howpublished = {\url{https://e2b.dev/}}
}

@inproceedings{agache2020firecracker,
  author    = {Alexandru Agache and Marc Brooker and Alexandra Iordache and Anthony Liguori and Rolf Neugebauer and Phil Piwonka and Diana-Maria Popa},
  title     = {Firecracker: Lightweight Virtualization for Serverless Applications},
  booktitle = {NSDI},
  year      = {2020}
}

@article{zhang2026prorlagent,
  author  = {Hao Zhang and Mingjie Liu and Shaokun Zhang and Songyang Han and Jian Hu and Zhenghui Jin and Yuchi Zhang and Shizhe Diao and Ximing Lu and Binfeng Xu and Zhiding Yu and Jan Kautz and Yi Dong},
  title   = {ProRL Agent: Rollout-as-a-Service for RL Training of Multi-Turn LLM Agents},
  journal = {arXiv preprint arXiv:2603.18815},
  year    = {2026}
}

@inproceedings{sheng2025hybridflow,
  author    = {Guangming Sheng and Chi Zhang and Zilingfeng Ye and Xibin Wu and Wang Zhang and Ru Zhang and Yanghua Peng and Haibin Lin and Chuan Wu},
  title     = {HybridFlow: A Flexible and Efficient RLHF Framework},
  booktitle = {EuroSys},
  year      = {2025}
}

@article{ni2026chimera,
  author  = {Kangqi Ni and Wenyue Hua and Xiaoxiang Shi and Jiang Guo and Shiyu Chang and Tianlong Chen},
  title   = {Chimera: Latency- and Performance-Aware Multi-agent Serving for Heterogeneous LLMs},
  journal = {arXiv preprint arXiv:2603.22206},
  year    = {2026}
}

@inproceedings{young2019gvisor,
  author    = {Ethan G. Young and Pengfei Zhu and Tyler Caraza-Harter and Andrea C. Arpaci-Dusseau and Remzi H. Arpaci-Dusseau},
  title     = {The True Cost of Containing: A {gVisor} Case Study},
  booktitle = {HotCloud},
  year      = {2019}
}

@misc{openai-sandbox-clients,
  author       = {{OpenAI}},
  title        = {Sandbox Clients in the OpenAI Agents SDK},
  year         = {2026},
  howpublished = {
    \url{https://openai.github.io/openai-agents-python/sandbox/clients/}
  }
}

@inproceedings{lagarcavilla2019software,
  author    = {H. Andr{\'e}s Lagar-Cavilla and Junwhan Ahn and Suleiman Souhlal and Neha Agarwal and Radoslaw Burny and Shakeel Butt and Jichuan Chang and Ashwin Chaugule and Nan Deng and Junaid Shahid and Greg Thelen and Kamil Adam Yurtsever and Yu Zhao and Parthasarathy Ranganathan},
  title     = {Software-Defined Far Memory in Warehouse-Scale Computers},
  booktitle = {ASPLOS},
  pages     = {317--330},
  year      = {2019}
}

@article{xu2026deepseek,
  title={Deepseek-v4: Towards highly efficient million-token context intelligence},
  author={Xu, Anyi and Lin, Bangcai and Xue, Bing and Wang, Bingxuan and Xu, Bingzheng and Wu, Bochao and Zhang, Bowei and Lin, Chaofan and Dong, Chen and Ling, Chenchen and others},
  journal={arXiv preprint arXiv:2606.19348},
  year={2026}
}

@misc{linux-userfaultfd,
  title        = {Userfaultfd},
  author       = {{Linux Kernel Documentation}},
  howpublished = {\url{https://docs.kernel.org/admin-guide/mm/userfaultfd.html}},
  note         = {Accessed: 2026-08-10}
}

@inproceedings{panwar2022tmcc,
  author    = {Gagandeep Panwar and Muhammad Laghari and David Bears and Yuqing Liu and Chandler Jearls and Esha Choukse and Kirk W. Cameron and Ali R. Butt and Xun Jian},
  title     = {Translation-Optimized Memory Compression for Capacity},
  booktitle = {MICRO},
  pages     = {992--1011},
  year      = {2022},
}

@article{jain2025r2e,
  title={R2e-gym: Procedural environments and hybrid verifiers for scaling open-weights swe agents},
  author={Jain, Naman and Singh, Jaskirat and Shetty, Manish and Zheng, Liang and Sen, Koushik and Stoica, Ion},
  journal={arXiv preprint arXiv:2504.07164},
  year={2025}
}

@article{wu2021practical,
  title={Practical temporal prefetching with compressed on-chip metadata},
  author={Wu, Hao and Nathella, Krishnendra and Pabst, Matthew and Sunwoo, Dam and Jain, Akanksha and Lin, Calvin},
  journal={IEEE Transactions on Computers},
  pages={2858--2871},
  year={2021}
}

@misc{zswap-lwn,
  author       = {Corbet, Jonathan},
  title        = {{The zswap compressed swap cache}},
  year         = {2013},
  howpublished = {\url{https://lwn.net/Articles/537422/}}
}

@misc{zswap-doc,
  title        = {{zswap -- The Linux Kernel Documentation}},
  year         = {2020},
  howpublished = {\url{https://www.kernel.org/doc/html/v5.8/vm/zswap.html}}
}

@inproceedings{wang2025openhands,
  title={Openhands: An open platform for ai software developers as generalist agents},
  author={Wang, Xingyao and Li, Boxuan and Song, Yufan and Xu, Frank F and Tang, Xiangru and Zhuge, Mingchen and Pan, Jiayi and Song, Yueqi and Li, Bowen and Singh, Jaskirat and others},
  booktitle={ICLR},
  volume={2025},
  pages={65882--65919},
  year={2025}
}

@article{yang2024swe,
  title={Swe-agent: Agent-computer interfaces enable automated software engineering},
  author={Yang, John and Jimenez, Carlos E and Wettig, Alexander and Lieret, Kilian and Yao, Shunyu and Narasimhan, Karthik and Press, Ofir},
  journal={NIPS},
  pages={50528--50652},
  year={2024}
}

@inproceedings{kumar2026tierscape,
  title={TierScape: Harnessing Multiple Compressed Tiers to Tame Server Memory TCO},
  author={Kumar, Sandeep and Prasad, Aravinda and Subramoney, Sreenivas},
  booktitle={EuroSys},
  pages={247--262},
  year={2026}
}

@inproceedings{jimenez2024swe,
  title={Swe-bench: Can language models resolve real-world github issues?},
  author={Jimenez, Carlos E and Yang, John and Wettig, Alexander and Yao, Shunyu and Pei, Kexin and Press, Ofir and Narasimhan, Karthik},
  booktitle={ICLR},
  pages={54107--54157},
  year={2024}
}

@inproceedings{lagar2019software,
  title={Software-defined far memory in warehouse-scale computers},
  author={Lagar-Cavilla, Andres and Ahn, Junwhan and Souhlal, Suleiman and Agarwal, Neha and Burny, Radoslaw and Butt, Shakeel and Chang, Jichuan and Chaugule, Ashwin and Deng, Nan and Shahid, Junaid and others},
  booktitle={ASPLOS},
  pages={317--330},
  year={2019}
}

@inproceedings{weiner2022tmo,
  title={TMO: Transparent memory offloading in datacenters},
  author={Weiner, Johannes and Agarwal, Niket and Schatzberg, Dan and Yang, Leon and Wang, Hao and Sanouillet, Blaise and Sharma, Bikash and Heo, Tejun and Jain, Mayank and Tang, Chunqiang and others},
  booktitle={ASPLOS},
  pages={609--621},
  year={2022}
}

@inproceedings{alecto,
  title={Integrating Prefetcher Selection with Dynamic Request Allocation Improves Prefetching Efficiency},
  author={Li, Mengming and Zhang, Qijun and Ren, Yongqing and Xie, Zhiyao},
  booktitle={HPCA},
  year={2025}
}

@inproceedings{prophet,
  title={Profile-Guided Temporal Prefetching},
  author={Li, Mengming and Zhang, Qijun and Gao, Yichuan and Fang, Wenji and Lu, Yao and Ren, Yongqing and Xie, Zhiyao},
  booktitle={ISCA},
  year={2025}
}

@inproceedings{kim1997stride,
  title={Stride-directed prefetching for secondary caches},
  author={Kim, Sunil and Veidenbaum, Alexander V},
  booktitle={ICPP},
  pages={314--321},
  year={1997}
}

@inproceedings{ainsworth2024triangel,
  title={Triangel: A High-Performance, Accurate, Timely On-Chip Temporal Prefetcher},
  author={Ainsworth, Sam and Mukhanov, Lev},
  booktitle={ISCA},
  year={2024}
}

@article{mittal2016survey,
  title={A survey of recent prefetching techniques for processor caches},
  author={Mittal, Sparsh},
  journal={CSUR},
  pages={1--35},
  year={2016}
}

@misc{linux-ksm,
  author       = {{Linux Kernel Documentation}},
  title        = {Kernel Samepage Merging},
  url          = {https://docs.kernel.org/admin-guide/mm/ksm.html},
  lastaccessed = {September 1, 2026}
}

@misc{linux-zram,
  author       = {{Linux Kernel Documentation}},
  title        = {zram: Compressed RAM-based Block Devices},
  howpublished = {\url{https://docs.kernel.org/admin-guide/blockdev/zram.html}},
  note         = {Accessed: 2026-08-25}
}

@misc{zeroboot,
  author       = {{Zeroboot}},
  title        = {Zeroboot: Sub-millisecond VM Sandboxes for AI Agents via Copy-on-Write Forking},
  year         = {2026},
  howpublished = {\url{https://github.com/zerobootdev/zeroboot}},
  note         = {Accessed: 2026-08-25}
}


\end{document}